## Title

**Learned Image resizing with efficient training (LRET) facilitates improved performance of large-scale digital histopathology image classification models**

## Authors
Md Zahangir Alom[1], Quynh T. Tran[1], Brent A. Orr[1]

## Affiliations
[1] Department of Pathology, St. Jude Children's Research Hospital

## Running title

**Financial support:** This work was partially supported by the National Cancer Institute support grant (P30 CA021765) and the American Lebanese Syrian Associated Charities (ALSAC).

## Corresponding Author
**Brent A. Orr MD, PhD, Department of Pathology, St. Jude Children's Research Hospital, 262 Danny Thomas Place, MS 250, Memphis, TN 38105-3678.**
**Email: brent.orr@stjude.org.Tel: (901) 595-3520; Fax: (901)**

## Conflict of interest

The authors declare no potential conflicts of interest.

**Word count:** 7496
**Figures:** 6
**Tables:** 1

**Abstract**

Histologic examination plays a crucial role in oncology research and diagnostics. The adoption of digital scanning of whole slide images (WSI) has created an opportunity to leverage deep learning-based image classification methods to enhance diagnosis and risk stratification. Technical limitations of current approaches to training deep convolutional neural networks (DCNN) result in suboptimal model performance and make training and deployment of comprehensive classification models unobtainable. Due to the massive size, each WSI must be partitioned into smaller patches prior to training or applying deep learning models. The input dimensions of DCNN architectures are small compared to the typical pathologist field of view, degrading performance by excluding important architectural features. Furthermore, data requirements for comprehensive models are sufficiently large to overwhelm the system memory during training.

In this study, we introduce a novel approach that addresses the main limitations of traditional histopathology classification model training. Our method, termed Learned Resizing with Efficient Training (LRET), couples efficient training techniques with image resizing to facilitate seamless integration of larger histology image patches into state-of-the-art classification models while preserving important structural information.

We used the LRET method coupled with two distinct resizing techniques to train three diverse histology image datasets, using multiple diverse DCNN architectures. Our findings demonstrate a significant enhancement in classification performance and training efficiency. Across the spectrum of experiments, LRET consistently outperforms existing methods, yielding substantial improvement of 15-28% in accuracy for a large scale, multiclass tumor classification task consisting of 74 distinct brain tumor types. LRET not only elevates classification accuracy but also substantially reduces training times, unlocking the potential for faster model development and iteration. The implications of this work extend to broader applications within medical imaging and beyond, where efficient integration of high-resolution images into deep learning pipelines is paramount for driving advancements in research and clinical practice.

**Introduction**

Histopathologic diagnosis is the gold standard for cancer diagnostics and forms the foundation of modern classification schemes. Although significant advances have been made in molecular testing, this information largely augments existing histopathologic classification with more refined molecular characterization. Traditionally, solid tumor tissue sections are stained with hematoxylin and eosin (H&E) and evaluated by a pathologist using light microscopy. Increasingly, H&E-stained sections are captured as whole slide images (WSI), and some centers have adopted fully digitized workflows [1,2].

Computational pathology is an emerging discipline that uses digital images to train models capable of adding clinical insights [3,4]. Most commonly, WSI labeled for class, or an alternative clinically relevant predictive variable, are used to train models such as deep convolutional neural network (DCNN) to predict the class labels [5]. This approach has been applied broadly in pathology across multiple organ systems [6]. Common variables predicted from standard H&E-stained sections include tumor type, genomic features of interest, and even outcome [7-11]. In combination with digital workflows, computational models have potential to transform the practice of surgical pathology through computer-aided diagnosis (CAD) [12,13].

Existing computational pathology models have major limitations that prevent effective implementation of CAD in the clinical environment. Most models are small-scale and non-comprehensive, restricted to only a few tumor classes, ranging between two to six tumor types [6]. While this may be suitable to specific use cases, implementation of models with so few classes severely limits more general clinical utility and creates a significant regulatory burden with respect to clinical implementation and validation. Small scale models are especially ineffective in organ systems such as the brain, in which the number of possible tumor types encountered may be greater than one hundred [14]. Implementation of large-scale comprehensive models, representing most tumor types encountered in a specific clinical practice would be ideal to minimize regulatory burden, minimize code and infrastructure maintenance, and to facilitate consideration across a broad range of differential diagnoses.

The substantial dataset size and massive size of individual WSIs both pose significant barriers to training and implementation of large-scale, comprehensive histopathology classification models. The overall dataset size requires training in batches to prevent system memory overload, and the load time cost during batch-by-batch learning can lead to prohibitively long training times. Because individual WSIs are too large to input directly to DCNNs, patch sampling is used to partition each WSI into thousands of compatible image patches, typically in the range of 224×224 pixels or 256×256 pixels. This approach yields thousands of patches per individual WSI. Patching images at full resolution yields images representing only focal tissue regions, excluding architectural features that are critical for optimizing histopathology classification performance [15,16]. In contrast resizing methods such as bilinear, bicubic, or nearest neighbor methods can convert larger images to sizes compatible with DCNN models, but with significant data loss and have been proven ineffective for training large, comprehensive models [13,17]. Talebi et al. proposed learned resizing (herein termed 'Google learned resizing (GLR)) networks to improve the efficiency of training computer vision tasks [18]. The GLR network incorporates traditional resizing methods within the end-to-end neural network during training to optimize the learned feature representation. Previous GLR has been applied to both classification, and image quality assessment, with the latter resulting in performance improvements over conventional methods [18]. While the proposed approach has potential application in computational pathology, the GLR or comparable resizing modules have not been applied to histopathology image analysis tasks.

In this study, we present novel methods designated learned resizing with efficient training (LRET) to facilitate training large, comprehensive histopathology classification models. We demonstrate that resizing modules, either GLR or our own novel resizing module designated High-dimension Feature Embedding (HFE), are compatible as input to diverse DCNN architectures using multiple large patch image sizes. Combined with efficient training methods, we demonstrate major improvements in histopathologic image classification performance while significantly reducing training time and resources requirements of training large multiclass models.

**Results**

***Development of learnable image resizing with efficient training (LRET) for large-scale histology image classification***

To optimize training for large scale histopathology image classification tasks, we needed to address two interconnected limitations of common training approaches. First, the size constraints imposed by the specified model input dimensions are significantly smaller than those utilized by pathologists for diagnostics, severely limiting the contribution of tissue architectural features in classification tasks. Second, the batch sizes needed to train across many classes can overwhelm the system memory. The latter problem is exacerbated by large input patch sizes. To address these issues, we developed a method designated learned image resizing with efficient training (LRET).

The first element of LRET is the learned image resizing (LR) modules, which we deployed directly upstream of the input layer of the respective DCNNs (Fig. 1a). Two distinct LR modules were explored, each learning a non-linear image representation while flexibly resizing an image patch to be compatible to the reduced input dimensions of the DCNNs. The first resizing block, designated Google learnable resizer (GLR), is a CNN based resizing method, like those described by Talebi et al [18]. The GLR module utilizes traditional resizing methods, such as bilinear or bicubic resizing, but connects the resizing with the DCNN model during training. This creates a machine compatible image manipulation but optimizes the lower dimension representations to emphasize the most informative features (Fig. 1b). The second resizing block, designated high-dimensional feature embedding (HFE) module, was constructed by placing two convolutional units in front of the DCNN (Fig. 1c). These convolutional blocks reduce the higher dimension input image to a lower dimensional embedding, compatible with the respective DCNN input dimensions. The HFE module maintains the relationship of the feature representation by sequentially reducing the feature dimensions to those of the DCNN model, while expanding the feature maps in z-dimension

(Fig. 1c). The HFE module is also placed in advance of the DCNN as part of training to emphasize the most important embedded representations.

The second component of LRET consists of method of efficient training (ET) large image models which significantly reduced the training time by sequentially queuing the training batches in cache and memory, similar to methods previously described for non-histology tasks [19]. Coupled with distributed methods allowing for use of multiple GPUs, large-scale ET significantly reduces the communication cost between data storage and memory, ensuring faster training.

***LRET training methods are flexible to diverse image input sizes.***

To validate the flexibility of LRET to fit histology patches from large input dimensions and train diverse computer vision models, we utilized two previously published histology datasets to train five DCNN models using LRET with GLR (LRET-GLR) or LRET with HFE (LRET-HFE). The first dataset represented a colorectal cancer (CRC) dataset consisting of three tumor and normal classes with input patch dimensions of 512×512 [9]. The second dataset, LC25000, consisted of a balanced set of five image classes representing a combination of normal and neoplastic colon or lung cancer with input patch dimensions of 768×768 [20]. These datasets were suited to evaluate the compatibility of the LRET training methods because they are well-annotated, the dimensions of 512×512 or 768×768 are larger than the input space of the selected models, and the classification performances have been previously established using reduced patch sizes [9,20].

The overall classification accuracy of LRET training on the CRC dataset using learned resizing of the 512 x 512 images outperformed the reported accuracy of >95% using the resized input dimension 256 x 256 [9]. In our analysis, DenseNet-121 trained using LRET-HFE showed the highest test accuracy (99.61%) with a weighted precision, recall, and F1-Score all greater than 0.99, respectively, despite utilizing the least number of learnable parameters (Fig. 2a and Supplementary Table 1). The ResNet50 model trained using LRET-GLR showed the worst performance on this dataset (99.55 %). Despite the observed variability, a range in performance improvement of 2.64 to 4.61% was observed for all models compared to the previous training methods using reduced input dimensions and traditional resizing methods [9]. These findings suggest that large patch sizes carry important classification information that may be lost in smaller patch images.

While the resizing modules uniformly improved classification performance, there was not a clear advantage to using the GLR module over the HFE resizing module. The Inception-v3 and EfficeintNet-B7 showed slightly better performance using the GLR module compared to the HFE module; however, the HFE module exhibited superior classification performance when coupled with the DenseNet-121 and XceptionNet models (Fig. 2 and Supplementary Table 1). A comparison of the training time and network parameters showed similarly mixed results, with no clear advantage of one resizing module over the other.

Confirming the flexibility of applying the LRET method to diverse input patch size, the LC25000 dataset, was trained using the default input patch sizes of 768×768 [20]. Like the findings in the

CRC dataset, we established clear compatibility of the LRET method to utilize diverse input patch dimensions and achieved performance improvement using selected resizing module-DCNN pairs (Supplementary Fig. 1 and Supplementary Table 2). The highest accuracy was observed with DenseNet121 models using learned resizing, with an accuracy of 99.62 and 99.4, for LRET-GLR and LRET-HFE trained models respectively. The best model, the DenseNet121-GLR, further surpassed 0.99 across the metrics of weighted precision, recall, and F1 score. This model's demonstrated an improvement in accuracy by 0.32% compared to the reported accuracy of 99.3% using standard input dimensions of 256×256 [20].

These findings suggest that LRET training is generally flexible for resizing large input patches and in many instances achieves model performances superior to traditional methods.

***LRET methods can scale to large multiclass classification datasets***
To test the scalability of our LRET methods to a large, comprehensive dataset we trained a publicly available 74-class brain tumor model using input patches approximating a pathologist's field of view (1024 x 1024) at 40 X magnification [21]. This brain tumor dataset (BTDB) consists of weakly annotated, non-overlapping patch images, and by the total number of unique classes, represents the largest existing published computational pathology dataset [6,21]. LRET- GLR or LRET-HFE methods were used to train the brain tumor set on multiple state of the art architectures, including Resnet50 [22], Inception-v3[23], XceptionNet [24], DenseNet-121[25], and EfficeintNet-B7 [26]. While models trained with either LRET-HFE or LRET-GLR each demonstrated significant performance improvement over the published validation accuracy of 66% [21], there was a trend toward superior accuracy achieved by the HFE resizing network (Fig. 3a and Table 1). The top performance was achieved by the LRET-HFE trained ResNet50 model, with a holdout test set accuracy of 94.16% and a weighted precision, recall, and F1-score of 0.943, 0.942, and 0.9416, respectively (Fig. 3b and Table 1 and Supplementary Fig. 2). In addition to overall high model accuracy of the LRET-HFE trained ResNet50 model, the majority of patches achieved high confidence class scores (90% > 0.9 compared to 1.77% in misclassified patches) (Fig. 3c). By applying stringent thresholding for high classification confidence, the patch level accuracy improved to 98.04 with a weighted F1 score of 0.98 on hold out image patches. These findings support the effectiveness of LRET methods to take very large patch sizes as input and significantly improves the performance of very large multiclass histopathology classification tasks using diverse DCNN architectures.

Notably, before adopting the ET approach, we previously used the BTDB to train the ResNet50 model using 280K total training patches at 1024×1024 pixels using the HFE module and traditional batch by batch learning. Although hold out testing 37K achieved 91.28% accuracy and a weighted precision, recall, F1-score of 0.9175, 0.9128, and 0.9130 respectively, the training time was infeasible for iterative model improvement (28 days on four NVIDIA P100-16GB nodes, 512 GB memory for 25 epochs), despite using fewer patch images for training. In comparison, the comparable model using 1.84× greater training set, completed 50 epochs of training on only two P100-16B nodes in 57.72 hours. These findings highlight the important contribution of efficient memory usage in the scalability of model training to large histopathology datasets.

Furthermore, to investigate the consequences of using the traditional resizing methods on testing accuracy for large-scale image classification tasks, we trained and tested a ResNet50 model with resized (256×256×3) patches, which yielded 71.89% testing accuracy on the same holdout test examples.

**Ablation Study results:**

A baseline ResNet50 model, designated ResNet50-1024, was also trained with original input patch dimensions (1024×1024×3) which represents the feature in very high dimensional space with enormous number of features mapped in the feature extraction layers. The resulting ResNet50-1024 model configuration requires more GPU memory to ensure uninterrupted training and required a significant increase in, size of the bottleneck layer features representation to 32×32×2048. This model was both computationally expensive and failed to achieve the performances of models trained with the learned resizing modules. For instance, the ResNet50-1024 showed 85.39% and 0.8554 testing accuracy and F1-score, respectively, in comparison to the 89.41% and 94.16% testing performance with GLR+ResNet50 and HFE+ResNet50 models. The improvement in accuracy of 4.02% and 8.77% compared to the baseline ResNet50-1024 model clearly illustrates the benefit of the proposed resizing modules.

There were also clear benefits of the resizing modules in respect to training time. The ResNet50-1024 model took 2.10 hours per epoch when we trained using two A100GPUs, on the other hand, the HFE-ResNet50 andGLR-ResNet50 trained models took only 32 minutes and 47 minutes per epoch, respectively. Hence, the GLR-ResNet50 and HFE-ResNet50 architectures are approximately 3.93 and 2.68 times faster, and computational the complexity is significantly lower than the ResNet50-1024 models without the learned resizing.

**LRET-HFE trained ResNet50 model is a powerful feature extractor.**

To gain further insight into the quality of the LRET-HFE trained ResNet50 model, we performed a deeper analysis of correctly and incorrectly classified patches from the hold out test set from the BTDB. First, because histology models can also be used to extract feature representations for downstream workflows [21,27], we evaluate the ability of the deep learned features to discriminate classes by performing t-stochastic neighbor embedding (t-SNE) of the 512-deep learned feature representation from the feature extraction layer of the LRET-HFE trained ResNet50 model. We observed overall good separation of classes in the unsupervised representation and regional separation of major tumor lineages (Fig. 4 and Supplementary Fig. 3). The reliability of the learned feature representations was supported by projection of the high-confidence test patches into the unsupervised embedding space of the training patches (Fig. 4).

To evaluate the histologic correlates of the model outputs, representative patch images were evaluated from true positive and false negative image patches (Fig. 5 and Supplementary Fig. 4). The high-confidence true positive set (>0.9 class score), representing 101,308 of holdout test patches, were characterized by high tumor content and preserved histomorphology (Fig. 5a-5h). A small proportion of cases (5.46 percent) were correctly classified, but with scores below 0.9. The associated patches from this 'low-confidence' true positive set were characterized by tissue artifacts and a high ratio of white space to tissue area (Supplementary Fig. 4). Most misclassified patches were associated with low confidence scores, and a high proportion represented misclassification within major tumor types or lineages (i.e. between meningioma histologic types or alternative grades of diffuse gliomas). We observed that the class prediction from some patches differed from ground truth but predicted a plausible alternative class. In select instances, these represent mislabeling of the weakly annotated patches (Supplementary Fig. 5).

One potential application of large histology models is to automate determination of tumor and normal tissue images from the larger WSI. Because tumor images commonly share architectural and cytomorphologic features compared to normal tissue, we explored the ability to utilize the additive classification scores from normal or tumor classes to discriminate tumor and normal tissues in holdout test images. The LRET-HFE trained ResNet50 model demonstrated good accuracy (0.99.46% percent), precision and recall for that task (0.99 each).

**LRET-HFE training is compatible with explainable AI methods**
To assess the compatibility of explainable artificial intelligence (AI) methods to our LRET training, we assessed the BTDB dataset LRET-HFE trained BTDB-ResNet50 model using GRAD-CAM [28], GRAD-CAM++ [29], and SCORE-CAM, three algorithms commonly used to assess pixel importance in classification tasks [30]. Conventional implementations of these algorithms in conjunction with the ResNet50 model using 256×256 input patches show the most accurate class-specific activation fidelity is represented in the 8×8-dimension feature representation. However, we found that the 32×32 feature representation of the LRET-HFE trained ResNet50 model was optimal for distinguishing tumor form non-tumor regions using GRAD-CAM (Fig. 6a). Extending this evaluation to selected tumor examples using additional explainable methods show class-specific activation maps enriched in the tumor compared to stromal regions (Fig. 6b) and suggests the LRET-HFE training method does not significantly alter the ability to implement explainable methods.

**Discussion**
In the current study we have developed LRET methods to resize histopathology images and efficiently train DCNN models. These methods flexibly accommodate patch image input sizes, including those approximating pathologist FOV, and are compatible with diverse DCNN architectures. Larger input patches preserve the histologic architectural features and enhance classification performance, while the improved efficiency allowed for quick iteration over multiple models even using the largest existing histopathology dataset. These methods facilitate training and deployment of large multiclass models that more closely approximate the needs of practicing pathologists.

The two major components of the LRET method, the learnable image resizing and the efficient training, have been separately introduced by Google Research for computer vision tasks [18]. The current study is the first to combine the two components and apply them to histopathology classification. In addition to the established LR methods, we have established a novel alternative resizing method, HFE, which learns a high-dimension feature embedding that is compatible to the DCNN input dimensions. The LRET methods overall appear to improve the training performance of DCNNs on histology data, leading to improvement over the published classification performance by at least one model on all datasets [9,21,31]. Notably, the performance enhancement was greatest for the large, multiclass brain tumor dataset which showed a range of improvements of 15-28% over the reported validation accuracy of 66% [21]. While holdout testing was not reported in the original publication, we found hold testing accuracy of 94.16% in our best models, and as high as 98.04% if performance is restricted to high-confidence (0.9) class calls.

While both resizing methods result in improved performance over conventional resizing methods, for large, multiclass models, the LRET-HFE training method appears to maintain the feature representations better than the GLR method. This is supported by the superior classification performance over diverse DCNN architectures. However, the LRET-GLR method may have an slight advantage in certain circumstances including when the training input dimension is not $2^n$ compatible (when n>8), as was evident in our training the LC25000 model which utilized an input dimension of 768×768. One additional advantage of using the LRET-HFE trained models is the compatibility with newer explainable methods including GRAD-CAM, GRAD-CAM++, and SCORE-CAM. Of note, we found that for 1024x1024 dimensional inputs, the optimal feature representation for the explainable methods is the 32 x 32-dimensional feature representation, in contrast to the 8 x 8 dimensional representation that is typically utilized for these respective methods [32].

Our methods could significantly impact the training and development of large scale, multiclass histology models. The majority of previous computational pathology studies are focused on few classes, typically containing three to four tumor classes [6]. Although we applied the method here to the largest histopathology image dataset of 74-classes, the methods could be utilized to train even larger models provided the appropriate memory and GPUs resources are available. Additionally, because we trained models using multiple state of the art architectures, the models trained as part of the current study could be used for transfer learning on other supervised tasks with flexibility to choose the model architecture for purpose. Outside supervised classification, the models established in the current study could show significant utility as feature extractors. Because of the diversity of tumor types included in the training data, the truncated feature extractor layers could be utilized for diverse unsupervised tasks including vector search or phenotypic cluster analysis [27,33]. This could have specific utility when applied to combined histology and spatial transcriptomics workflows, for instance, as was recently proposed by Dent et al. [27].

The LRET method has potential for future optimization over our initial implementation. For instance, we have utilized patch images up to 1024 × 1024 pixels as this represents a typical pathologist FOV; however, it isn't clear if there is an upper bound for which the increase in patch size degrades performance on large multiclass histopathology datasets. Because the BTDB training set came from extracted patches rather than original WSI, alternative datasets would be necessary to further explore optimized patch size dimensions. It will also be important to evaluate the generalizability of the respective LRET trained models outside the original dataset. The brain tumor set is labeled by patch but is not distinguished at the slide level.  Although the patches were independent in the training and testing sets, patches were assigned randomly from the set of class-specific patches and therefore patches from the same case could be present in both sets. This constraint may overestimate the test performance compared to what might be obtained from independent slides. To determine the true generalizability of the model, additional testing will need to be performed on tumors from a completely independent dataset, preferably from WSI images.

In conclusion, we have developed LRET methods to facilitate efficient train large DCNN model with very high classification accuracy. We demonstrate performance improvements of up to 28% improvement in accuracy in the largest WSI dataset.  Future application of this method to other large comprehensive datasets should make clinical tumor classification from large comprehensive tumor datasets feasible.

**Methods**

**Image dataset preparation**

The training, validation, and testing of the DCNN models were performed using three different publicly available datasets for colon cancer [9,34], combined lung and colon cancer [20], and brain tumor classification tasks [21]. Images from the respective datasets were used directly for model training and testing without filtering on blank space.

*Colorectal cancer (CRC) dataset*:  The CRC dataset [9,34] was downloaded from [https://zenodo.org/records/2530789]. The dataset consists of 11,977 total number of samples and three classes: adipose and mucinous tissue (ADIMUC), stroma and muscle tissue (STRMUS), and tumor epithelial tissue (TUMSTU).  The base dimensions of patch images are 512×512×3, which were used directly as input to the resizing modules during training and testing. The dataset was used without further preprocessing as released [34]. From the entire set of image patches, 80 percent were randomly allocated to training and 20 percent to testing (10,180 patches and 1797 patches, respectively). All available patch images were considered for modeling without under sampling to balance class.

*Lung and Colon (LC25000) datasets*:  The LC25000 dataset was downloaded from [https://academictorrents.com/details/7a638ed187a6180fd6e464b3666a6ea0499af4af] [20]. The dataset contains 25,000 total images with base patch input dimensions of 768×768×3 pixels [20]. The dataset is balanced, consisting of 5000 images from each of five tumors or benign tissue classes, including colon adenocarcinoma, benign colonic tissue, lung adenocarcinoma, lung squamous cell carcinoma, and benign lung tissue.   For model development, images were split proportionally by class with 80 percent of the patch images (20K) used in the training set and the remaining 20 % (5000 images) used in the independent testing set.  Twenty percent of the images from the training set (4000 images) were used for validating the model during the training process.

*Brain tumor dataset* (BTD): The brain tumor dataset was downloaded from [https://doi.org/10.5281/zenodo.3234829] [21]. The dataset consisted of 838,644 weakly annotated and extracted patch tiles from diverse tumors, non-neoplastic, and tissue artifacts, representing 74 different histologic classes [21]. The base dimension of the patch images is 1024×1024 pixels. The dataset demonstrates significant class imbalance, ranging from 483 to 74,576 images per class. To minimize the effects of class imbalance, we placed an upper bound of 20,000 patches per class. In total, 676,544 images were utilized, partitioned into training and test sets at a ratio of 85:15 as was previously reported [21]. Weighted class training was performed using the Sklearn-package to minimize the effect of class imbalance during training. Although some image patches contained significant areas devoid of tissue, no tiles were discarded with respect to the blank regions.

**DCNN model and optimization:**

End-to-end classification architectures were constructed for large-scale histopathological image classification tasks using two learnable resizers modules including the proposed higher dimension feature embedding (HFE) resizing module and the Google Learnable Resizer (GLR) [18].

**High-dimension feature embedding (HFE):** The HFE module was developed to down sample the input patches using different convolutional units, embedding the input patches into a high dimension feature embedding space. The first convolutional unit constructed with two convolutional layers with kernel size of 3×3 and strides of 1 and 2, respectively. The number of kernels in the first unit was set to 4. Then, a batch normalization layer was applied, followed by an activation layer where a Rectified Linear Unit (ReLU) activation function was used. Next, a max pooling layer was used with a 3x3 kernel where a stride size of 2×2 was applied, prior to input to the second convolutional unit. The second convolutional unit consists of two convolutional layers with a kernel size of 3x3 and the number of kernels set to 8. The final output of the HFE module characterized by dimensions of 256×256×8, representing the higher dimension features embedding. The outputs of the HFE are subsequently used as inputs to the baseline classification models. To accommodate eight input channels to the classification model, the input channels of the baseline classifier was updated and set to eight. For input size of 1024x 1024x3, we applied both convolutional units with sub-sampling layers, on the other hand, we skipped sub-sampling layer for 512×2512×3 and 768×768×3 dimensional input patches, respectively.

**Google Learned Resizing Module (GLR):** the GLR developed by Google was utilized in the initial layers of models to resize to the higher dimensional input patches to the lower dimension input space of 224×224×3, and 299×299×3, which are appropriate for the DCNN classifiers. The GLR was constructed in combination with a bilinear features resizer, residual blocks, and skip connection as previously described [18].

**End to end DCNN model construction:** In this study, five different baseline DCNN classifiers including ResNet50 [22], Inception-v3[23], XceptionNet [24], DenseNet121 [25], and EfficientNet-B7 [26], were used to construct classification models trained and tested on three different datasets. All baseline DCNN models were trained from scratch without using pretrained ImageNet weights. For the HFE-based classifiers, the input dimensions of the baseline DCNN models were set to 256×256×8. On the other hand, for the GLR-based models, the input dimensions were set to 224×224×3 for all DCNN architectures, except the Inception-v3 which utilized 299×299×3 dimensional inputs. The input resizing module (HFE or GLR) was coupled with the baseline DCNN models without top layers which was named the feature extraction network (FEN). The dimension of FEN outputs is 8x8x512. To reduce the learnable parameters and speed up the training process, the fully connected layers of the baseline models were replaced with a Global Average Pooling (GAP) layer, embedding the representation to a 512-dimensional space [35]. Then, a layer with 512 neurons was used followed by a dropout layer with a dropout probability rate of 0.5. Finally, a SoftMax classification layer was used at the end of the model to compute outputs confidence probability. The end-to-end architecture, including the HFE or GLR module, the modified baseline classifiers, and top classification layers were jointly trained to optimize the models' parameters. Furthermore, to investigate the impact of the individual resizing modules on the performance of the models, we evaluated the models under conditions of ablation.

**DCNN optimization:**

The efficient training was done using a recently developed data loading pipeline methods from Google TensorFlow [36]. During the training process, we used cache () to supply the input images in memory after loading off the inputs from the disk starting with the first epoch, and buffered prefetching was used to allow later samples to be loaded while the current samples are being processed during training. As a result, this data loading pipeline helps reduce the data loading time significantly and ensures efficient high throughput training. In addition, to ensure faster training using multiple GPUs, we utilized the TensorFlow distributed data parallelism strategies released from Google [36].

*Optimizers:* For the CRC and *LC25000* datasets, the models were trained for 100 epochs using a batch size of 32 for EfficeintNet-B7 and 64, was used for the rest of the models. The BTDB, was trained for 50 epochs with a batch size of 16 and 32. The models were trained using the Adam optimizer with default parameters ($\beta_1 = 0.9$, $\beta_2 = 0.999$, and learning rate$(lr) = 0.001$) [37]. The validation accuracy was monitored to save the best model during the training phase. Then, the best model from the respective model build was used for testing independent test examples.

**Model performance evaluation metrics:** Scikit-learn was used to calculate classification performance metrics of overall accuracy, weighted precision, weighted recall, and weighted F1-score [38]. Also, the confusion matrices and receiver operating characteristics (ROC) with area under the curve (AUC), and precision-recall curves (PRC) were generated to visualize the performance of classification models.

**Large-Scale Ablation Study**

To investigate the impact of the individual resizing module (HFE or GLR), we conducted a set of experiments in which the resizing modules were excluded from the end-to-end architecture. We chose the best ResNet50 model trained and tested using the same training and testing example sets from the brain tumor dataset. A baseline ResNet50 model with identical top layers was trained with original input patches (1024×1024×3), as a result, the size of the bottleneck layer features representation was 32×32× 2048. The performance metrics of the base model without LR was compared to the respective models with the LR modules.

**Deep histological feature representation and clustering:**

For brain tumor classification tasks, the deep-learned features were extracted from bottleneck layer of the LRET-HFE-ResNet50 model which showed the best testing performance. The dimension of the feature representation was 8×8×512. To generate weighted feature representation with respect to the classes, we randomly selected the representation from both training and testing features and performed averaging. We explored different sampling sizes including 2,3,5, and 10 feature representations from each class. In contrast, the previous study reported the visualization of feature representation averaging by 20 [21]. However, we found averaging by three feature representations per class not only secured better representation but also preserved the intra and inter-class tumor relationships. Hence, the final deep-learned feature vectors (DFV) were generated from training and testing examples by averaging the representation by 3. Next, t-stochastic neighbor embedding (t-SNE) [39] was applied on the 512 deep learned feature representation space (averaged by 3) with the following parameters (verbose=1, perplexity=40, number of iterations=300). The t-SNE coordinates were visualized with respect to their respective class label.

**Explainable AI (XAI):**

Deep learned feature interpretability was preformed using Gradient-weighted Class Activation Mapping (Grad-CAM), Gradient-weighted Class Activation Mapping ++ (Grad-CAM++), and Score-weighted visual explanations for convolutional neural nets (SCORE-CAM) [28-30]. Briefly, the Deep learned feature representation from the final layer of the LRET-HFE trained ResNet50 model were extracted from the independent test samples. Then, the Grad-CAM [28], Grad-CAM++ [29], and SCORE-CAM [30] methods were applied to generate the heatmap representation for visualization on feature representation layers for 8×8, 16×16, 32×32, and 64×64.

Both Grad-CAM and Grad-CAM++ are gradient-based methods, where Grad-CAM generates visualization based on a CNN model with fine-grained details of the predicted class [28]. The Grad-CAM++, a generalized visualization approach with pixel-wise weighting of the gradients of the outputs with respect to the special position of the last layer of FEN [29]. On the other hand, the SCORE-CAM, a novel gradient-free visual explanation model has proposed in 2020 by combining perturbation and CAM [30]. In this study, we chose the best LRET-HFE trained ResNet50 model and generated the deep learned feature representation from the last layer of FEN for independent test examples. Then, the Grad-CAM [28], Grad-CAM++ [29], and SCORE-CAM [30] methods were applied to generate the heatmap representation for visualization.

**Components:**

**Software, packages, and hardware used:**

Here is the list of the packages, we used for implementing the proposed model in python 3.8.15 environment: TensorFlow 2.8 [36], python computer-vision library (CV2 4.7.0) [40], Pandas(1.5.2) [41], Scikit-learn (1.2.1) [38].

Regarding the hardware, the models were trained on St. Jude HPC-GPU cluster with 2-GPU or 4-GPU NVIDIA A100-SXM-80GB and NVIDIA P100-PCIE-16GB based on the availability of the resources. However, the computational time reported in this study with respect to 2-P100 GPUs.

**Reporting summary:**
Additional summary on research design is available.

**Data availability:**
All datasets used and analyzed in this study are publicly available: for colon cancer dataset, visit: [9]. For colon and lung cancer datasets, visit: [20] [31]. And for the brain tumor datasets, please visit the following link: [21]

**Code availability:**
The code for all our model implementations is publicly available at the following link: [link to be added upon publication]

**Acknowledgements**
This work was partially supported by the National Cancer Institute support Grant (P30

CA021765) and the American Lebanese Syrian Associated Charities (ALSAC).

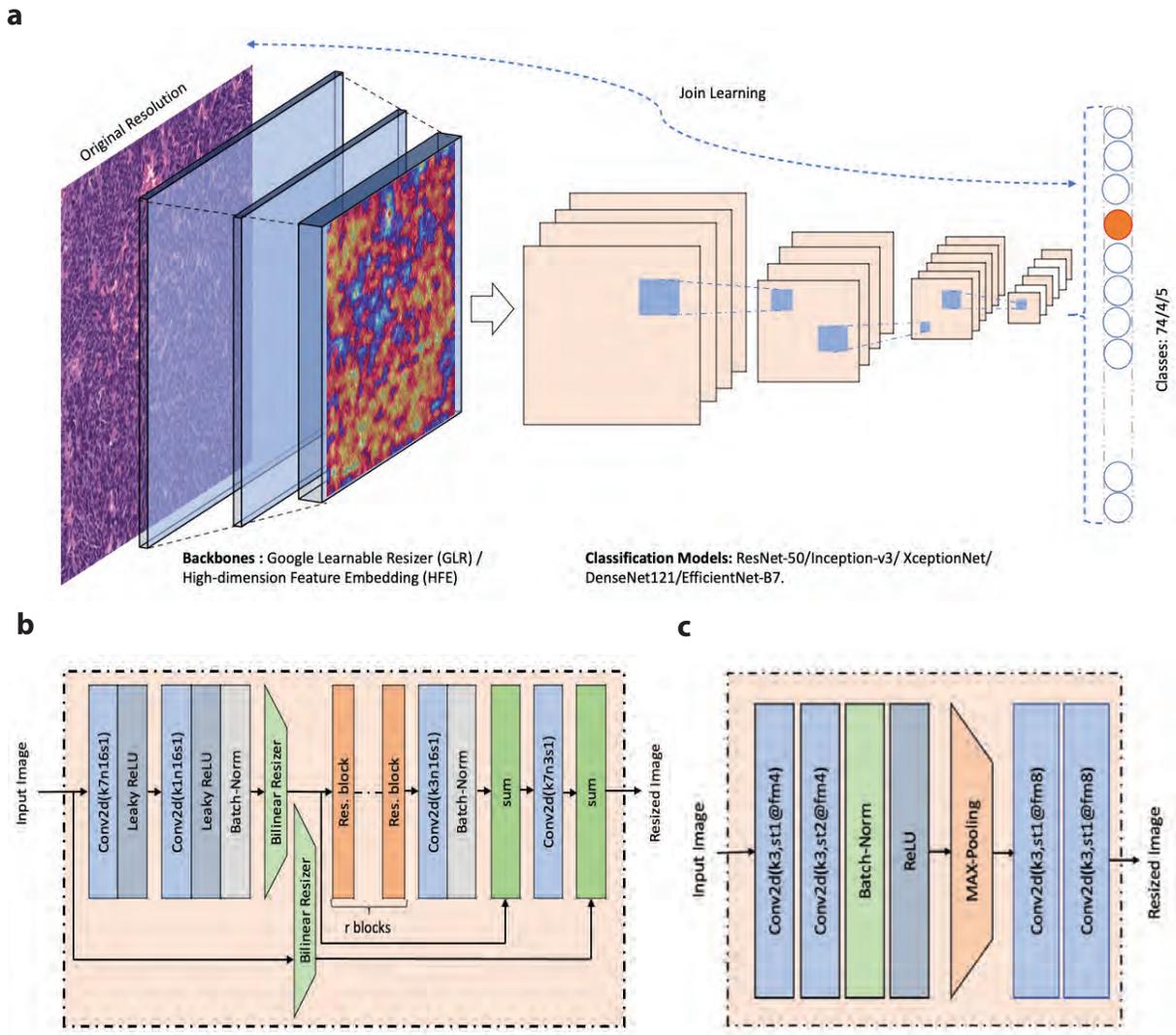

**Figure 1. End-to-end architecture for LRET trained classifiers. a.** The LRET-trained classification models consist of two primary components, the resizing module, and the baseline classifier with customized classification layers. Two separate resizing modules were evaluated in this study, including the learned resizing **b.** Google learned resizing (GLR) module and the **c.** high-dimension feature embedding (HFE) module. The complete LRET-training method also reduces training time by employing end-to-end efficient training using a cache-centric approach.

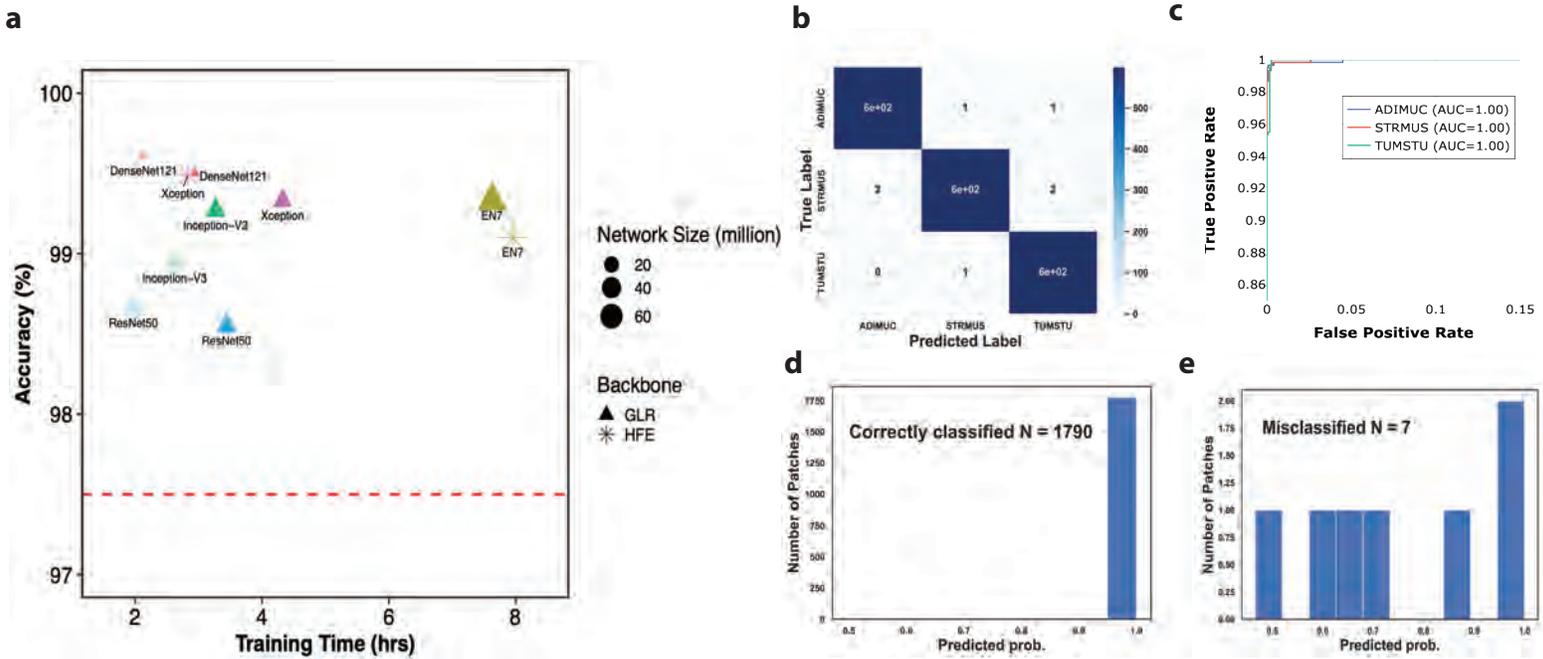

**Figure 2. Validation of LRET training using diverse DCNN architectures and 512 x 512 image patches.** LRET methods were used to train models from a colon cancer dataset consisting of three balanced classes of tumor epithelial tissue (TUMSTU), mucinous tissue (ADIMUC), and stroma and muscle tissue (STRMUS) using starting patch images of 512 x 512 9,34. Diverse DCNN architectures were trained using either the GLR or HFE resizing module including, ResNet50, Inception-v3, XceptionNet, DenseNet121, and EfficeintNet-B7. **a.** The training time versus accuracy is presented for each resizing module-DCNN pair. Models trained using the GLR resizing module are depicted as triangles and models trained using the HFE module are depicted as stars. The symbol size is proportional to the number of model parameters. The reported accuracy for the model trained using conventional methods is presented as a red hatched line. **b.** The confusion matrix and the **c.** Receiver operator curve (ROC) for the hold out test samples. The class score probability distributions for **d.** correctly classified and e. misclassified samples, respectively.

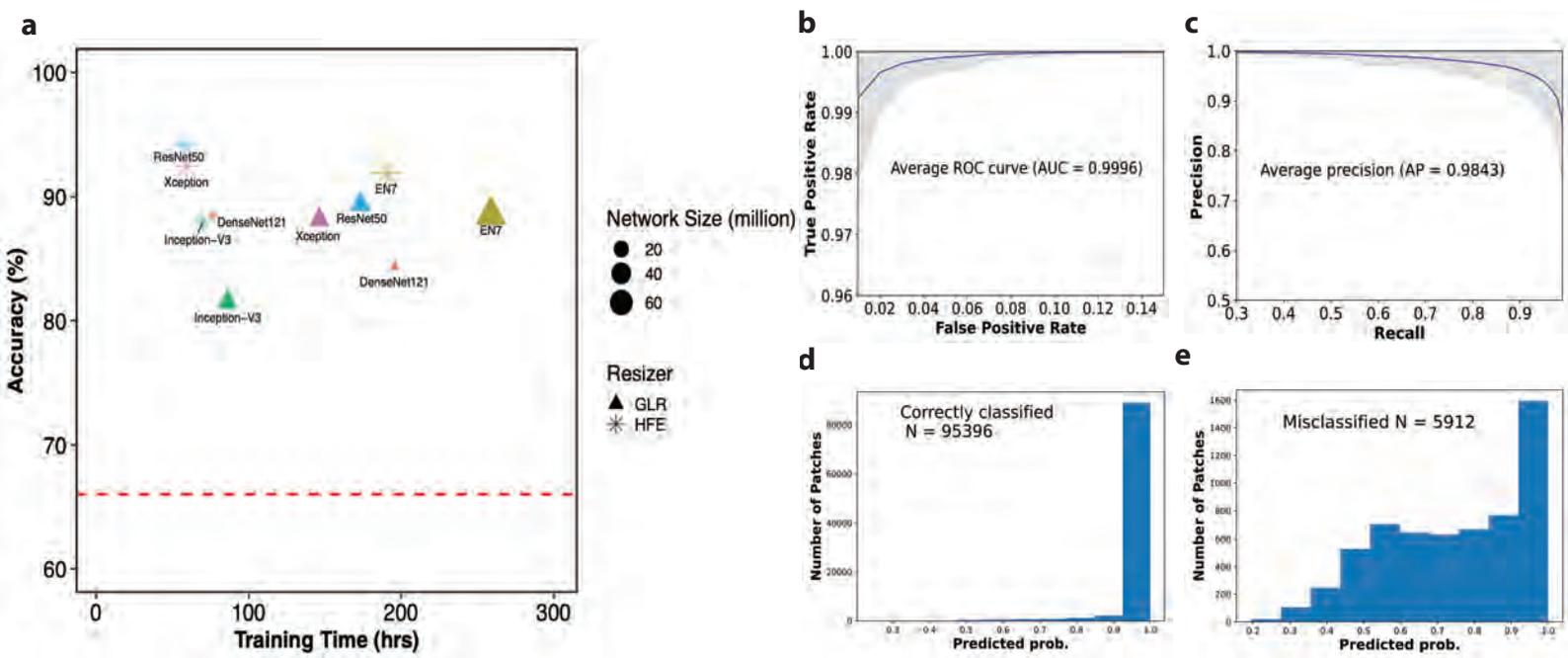

**Figure 3. Validation of LRET training on large comprehensive brain tumor dataset.** a. LRET methods were used to train a 74 class brain tumor dataset using field-of-view input dimensions of 1024 x 1024. The overall test accuracy compared to the training time in hours is plotted for ResNet50, Inception-v3, XceptionNet, DenseNet121, EfficeintNet-B7 using GLR (triangles) or HFE (stars) resizing modules. Models trained using the GLR resizing module are depicted as triangles and models trained using the HFE module are depicted as stars. The symbol size is proportional to the number of model parameters. The red hatched line represents the previously reported validation accuracy of 66% for this dataset. **b.** The average ROC with Area under curve (AUC = 0.9996) on hold-out test images for the LRET-HFE trained ResNet50. **c.** Precision versus recall curve for the for the LRET-HFE trained ResNet50 model on hold out testing with average precision (AP = 0.9843). **d.** The LRET-HFE-ResNet50 model output confidence value distribution for the correctly classified patches. **e.** The output confidence value distribution for the misclassified patches.

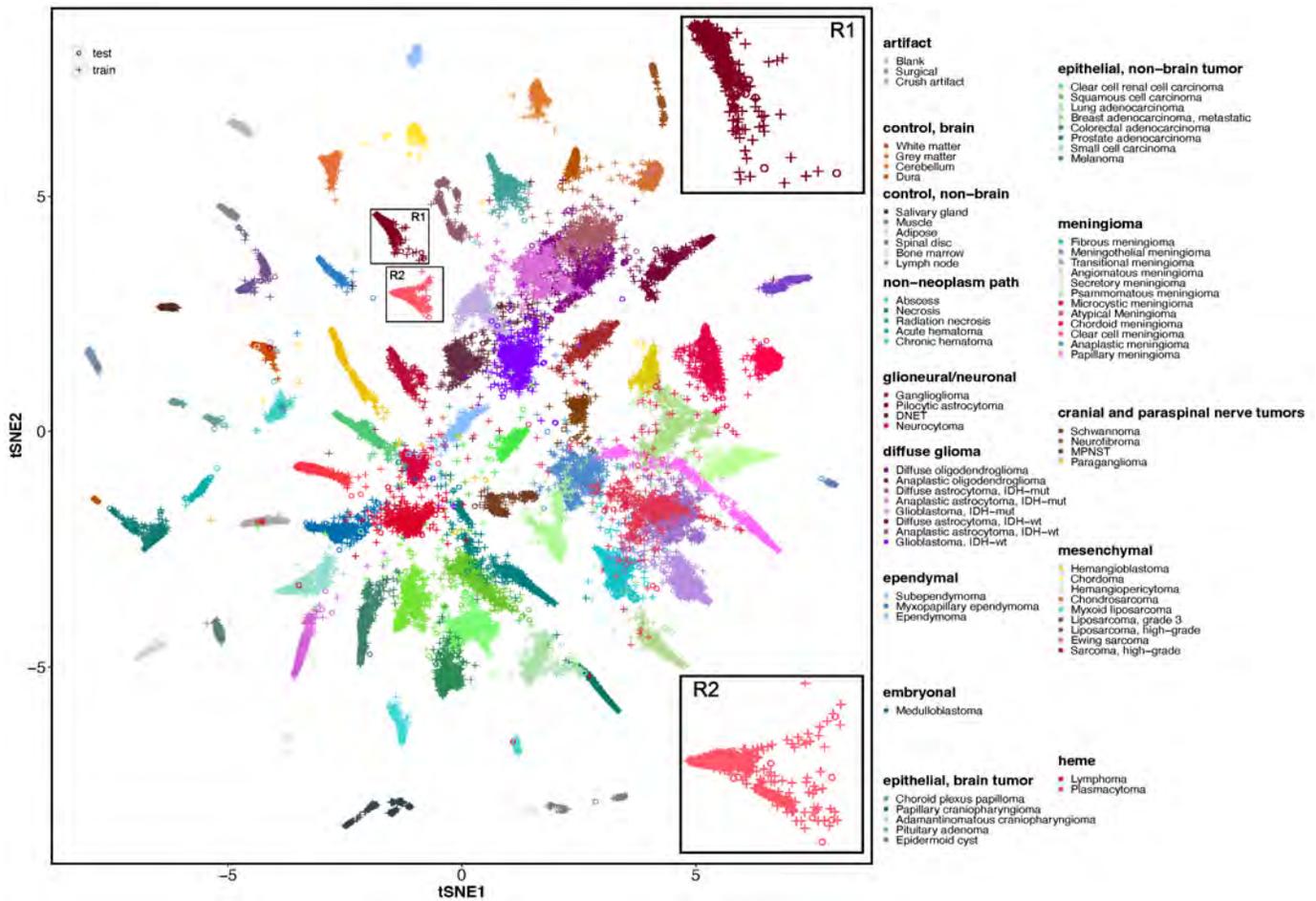

**Figure 4. t-SNE projection of latent space representation of training and test images from LRET-HFE trained Resnet50 model.** The 512 deep learned feature representation is plotted using t-stochastic neighbor embedding for the training data (plus sign) and test data (open circles) and colored by the tumor class. The plot demonstrates good separation of the latent space representations by tumor type and the test samples project into the same feature space as the training data. The overlap of the training and test data are highlighted in the zoomed in images for pilocytic astrocytoma (R1) and clear cell meningioma (R2).

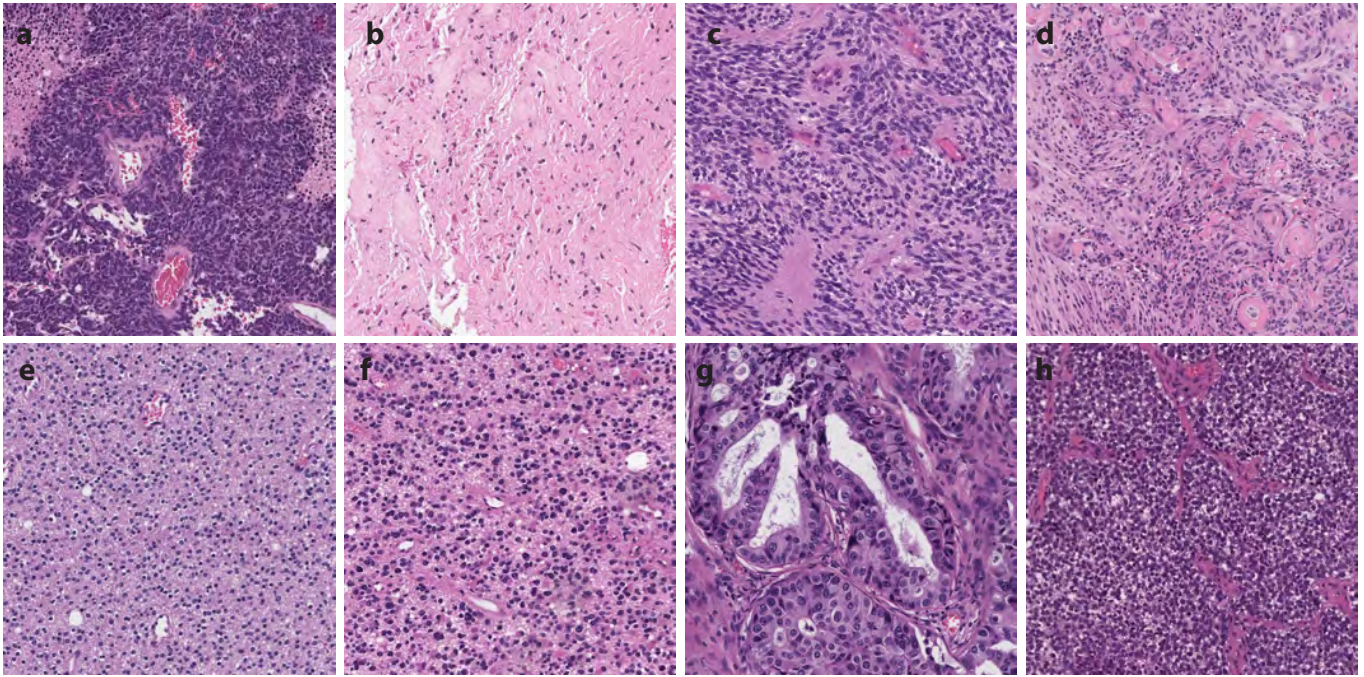

**Figure 5. Evaluation of high-confidence patch images from LRET-HFE trained ResNet50 brain tumor model.** Selected 1024 x 1024, H&E-stained hold-out test patch images with high-confidence classification scores (class score > 0.9) representing true positive class calls for **a.** medulloblastoma, **b.** pilocytic astrocytoma, **c.** glioblastoma, IDH-wildtype, **d.** meningothelial meningioma, **e.** oligodendroglioma, **f.** diffuse astrocytoma, **g.** metastatic lung adenocarcinoma, and **h.** Ewing sarcoma.

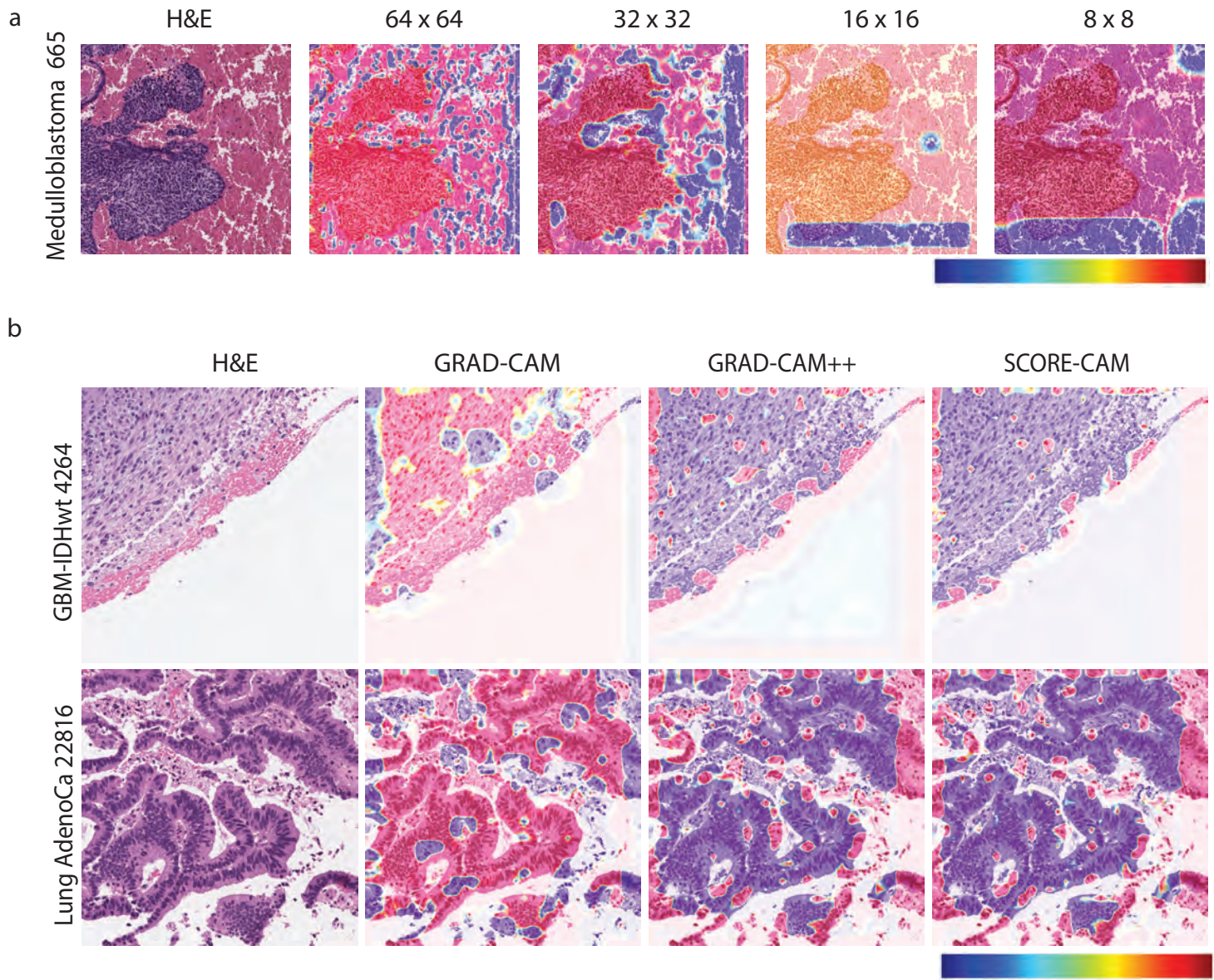

**Figure 6. Implementation of explainable methods on LRET-HFE trained ResNet50 model.**
**a.** Representative H&E patch image and heatmap representation of pixel importance at the different representation dimensions demonstrates the 32 × 32 is optimal for the LRET-HFE trained ResNet50 model using the GRAD-CAM algorithm. **b.** Representative H&E patch images and importance score heatmap representation from glioblastoma, IDH-wildtype (GBM-IDHwt) and metastatic lung adenocarcinoma (Lung AdenoCa) using diverse explainable methods including GRAD-CAM, GRAD-CAM++, and SCORE-CAM. The pixel importance ranges from blue (least important) to red (most important).

| LR module | Models | Precision | Recall | F1-score | Accuracy |
|---|---|---|---|---|---|
| GLR | DenseNet121 | 0.8620 | 0.8437 | 0.8451 | 84.37 |
| GLR | ResNet50 | 0.9008 | 0.8941 | 0.8945 | 89.41 |
| GLR | EN7 | 0.8969 | 0.8854 | 0.8865 | 88.54 |
| GLR | Xception | 0.8905 | 0.8817 | 0.8819 | 88.17 |
| GLR | Inception-V3 | 0.8304 | 0.8151 | 0.8146 | 81.51 |
| HFE | DenseNet121 | 0.8966 | 0.8852 | 0.8852 | 88.52 |
| **HFE** | **ResNet50** | **0.9427** | **0.9416** | **0.9416** | **94.16** |
| HFE | EN7 | 0.9218 | 0.9191 | 0.9192 | 91.91 |
| HFE | Xception | 0.929 | 0.9243 | 0.925 | 92.43 |
| HFE | Inception-V3 | 0.8913 | 0.8805 | 0.8808 | 88.05 |

**Table 1. Performance metrics for models trained on the brain tumor dataset using LRET**. Multiple DCNN models were trained on seventy-four class brain tumor dataset using the GLR or HFE resizing module. The weighted precision, weighted recall, weighted F1-score, and accuracy were calculated on hold-out test images. The best performance for each metric is represented by bold text.

| LR module | Models | Precision | Recall | F1-score | Accuracy |
|---|---|---|---|---|---|
| GLR | DenseNet121 | 0.9949 | 0.99499 | 0.9949 | 99.49 |
| GLR | ResNet50 | 0.9855 | 0.9855 | 0.9855 | 98.55 |
| GLR | EN7 | 0.9933 | 0.9933 | 0.9933 | 99.33 |
| GLR | Xception | 0.9933 | 0.9933 | 0.99332 | 99.33 |
| GLR | Inception-V3 | 0.9927 | 0.99276 | 0.99276 | 99.27 |
| **HFE** | **DenseNet121** | **0.9961** | **0.9961** | **0.9961** | **99.61** |
| HFE | ResNet50 | 0.9868 | 0.98664 | 0.9866 | 98.66 |
| HFE | EN7 | 0.9911 | 0.991 | 0.9911 | 99.10 |
| HFE | Xception | 0.9949 | 0.9949 | 0.9949 | 99.49 |
| HFE | Inception-V3 | 0.9894 | 0.9894 | 0.9894 | 98.94 |

**Supplementary Table 1. Performance metrics for models trained on Colorectal cancer (CRC) dataset using LRET**. Multiple DCNN models were trained on CRC dataset using the GLR or HFE resizing module. The weighted precision, weighted recall, weighted F1-score, and accuracy were calculated on hold-out test images. The best performance for each metric is represented by bold text.

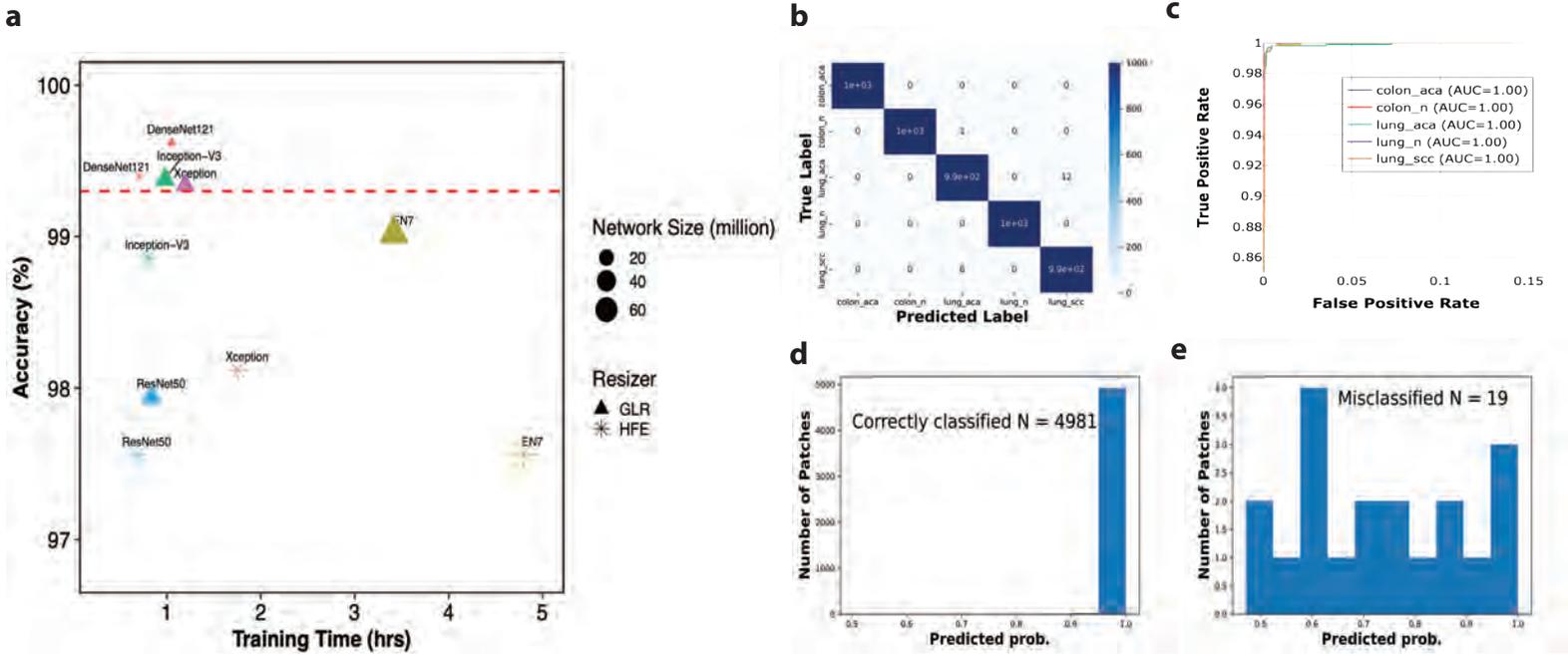

**Supplementary Fig. 1. Validation of LRET training using diverse DCNN architectures and 768 x 768 image patches.** LRET methods were used to train models from a colon cancer dataset consisting of five classes, including colon adenocarcinoma (colonic_aca), benign colonic tissue (colon_n), lung adenocarcinoma (lung_aca), lung squamous cell carcinoma (lung_scc), and benign lung tissue (lung_n) using patch images of 768 x 768 20. a. The training time versus accuracy is presented for each resizing module-DCNN pair using diverse DCNN architectures and the GLR (triangles) or HFE (stars) resizing modules. The symbol size is proportional to the number of model parameters. The reported accuracy for the model trained using conventional methods is presented as a red hatched line. b. The confusion matrix for the hold out test samples is presented for the DenseNet121-GLR trained model. c. The ROC curve is for the DenseNet121-GLR model. d. For the DenseNet121-GLR The class score distributions for correctly classified and e. misclassified samples are presented, respectively.

| LR module | Models | Precision | Recall | F1-score | Accuracy |
|---|---|---|---|---|---|
| **GLR** | **DenseNet121** | **0.9962** | **0.9962** | **0.9962** | **99.62** |
| GLR | ResNet50 | 0.9796 | 0.9794 | 0.9794 | 97.94 |
| GLR | EN7 | 0.9902 | 0.9902 | 0.99019 | 99.02 |
| GLR | Xception | 0.9934 | 0.9934 | 0.9933 | 99.34 |
| GLR | Inception-V3 | 0.9938 | 0.9938 | 0.9937 | 99.38 |
| HFE | DenseNet121 | 0.994 | 0.994 | 0.9939 | 99.40 |
| HFE | ResNet50 | 0.9759 | 0.9756 | 0.9755 | 97.56 |
| HFE | EN7 | 0.9762 | 0.9756 | 0.9755 | 97.56 |
| HFE | Xception | 0.9817 | 0.9812 | 0.9811 | 98.12 |
| HFE | Inception-V3 | 0.9886 | 0.9886 | 0.9885 | 98.86 |

**Supplementary Table 2. Performance metrics for models trained on the Lung and Colon cancer (LC25000) dataset using LRET.** Multiple DCNN models were trained on LC25000 dataset using the GLR or HFE resizing module. The weighted precision, weighted recall, weighted F1-score, and accuracy were calculated on hold-out test images. The best performance for each metric is represented by bold text.

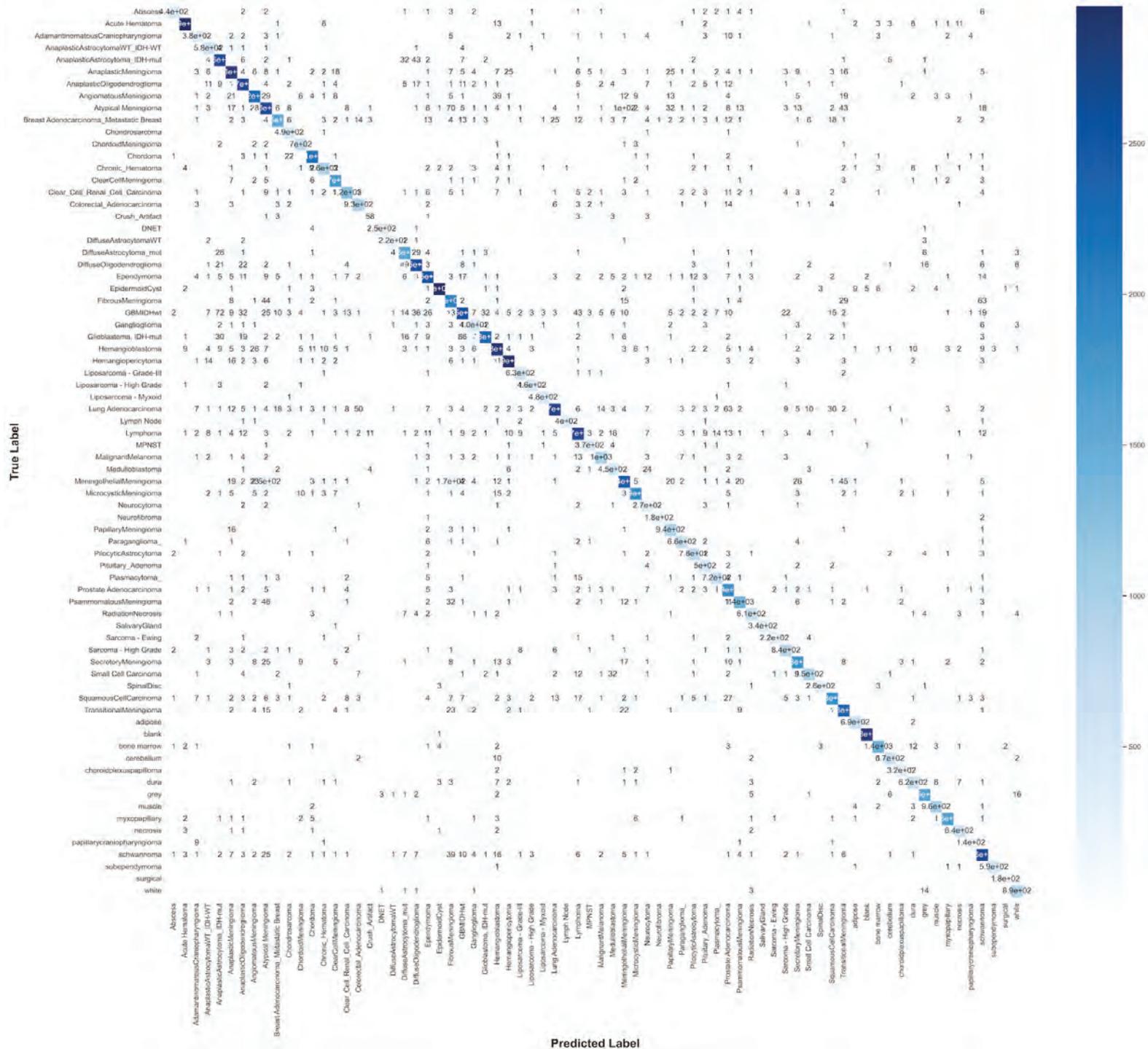

**Supplementary Fig. 2. The confusion matrix for 74 different brain tumor classification tasks for the independent test examples.** The LRET-HFE trained ResNet50 model was used to classify hold-out test images. The confusion matrix is displayed with true labels on the x-axis and predicted labels on the y-axis. The proportion of examples falling in each true and predicted pair is depicted from white (low proportion) to dark blue (high proportion).

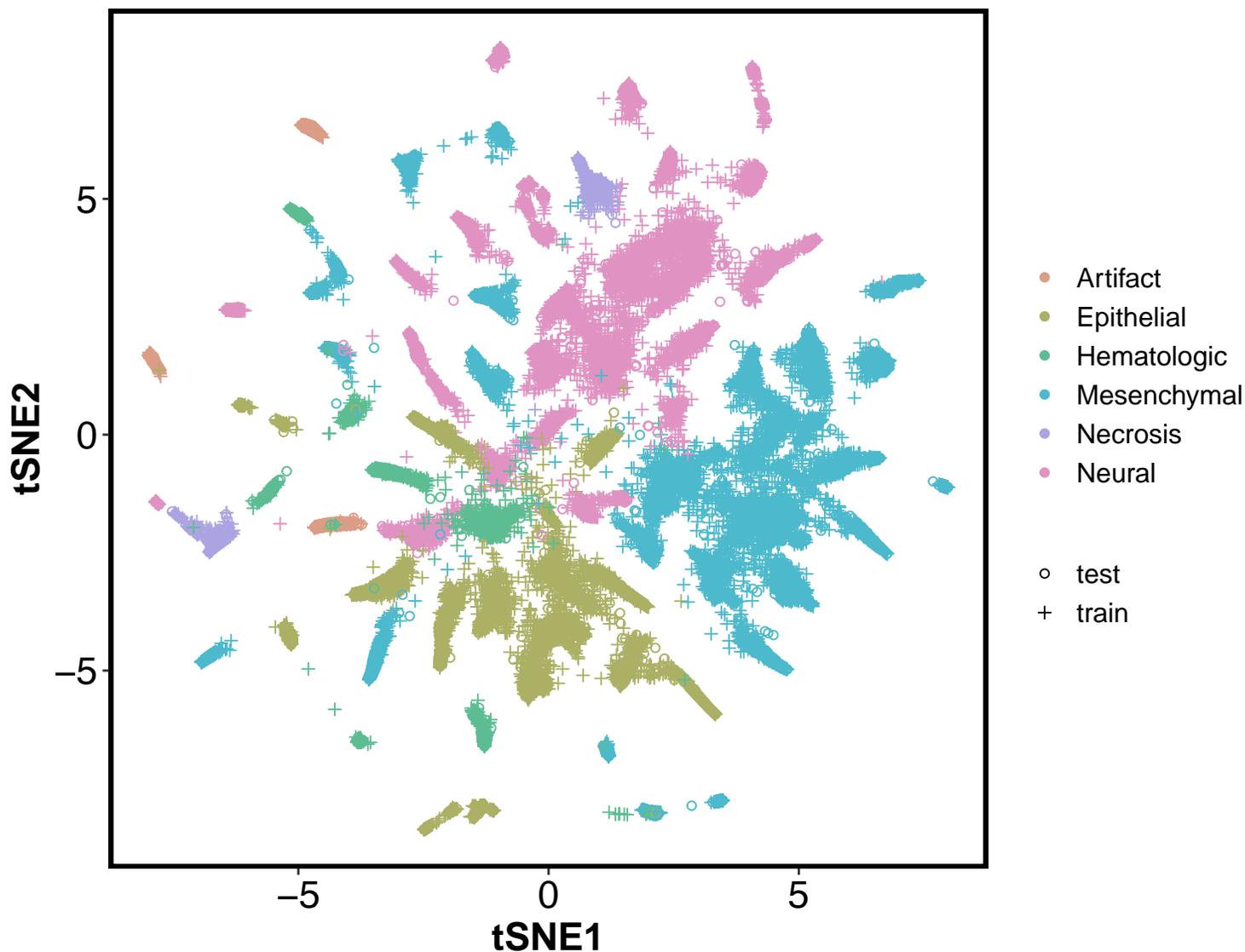

**Supplementary Fig. 3. The t-SNE plot for group level analysis for 74 different brain tumors classification.** The t-SNE image from Figure 4 is colored by selected tumor and normal lineage groups including artifact, epithelial tissue, hematologic tissue, mesenchymal tissue, neural tissue, or necrosis. The projection demonstrates clear distinction of the lower dimension projection of deep learned feature representation of these lineages based on morphology. The training examples are depicted as a plus symbol and the test example images are represented as circles.

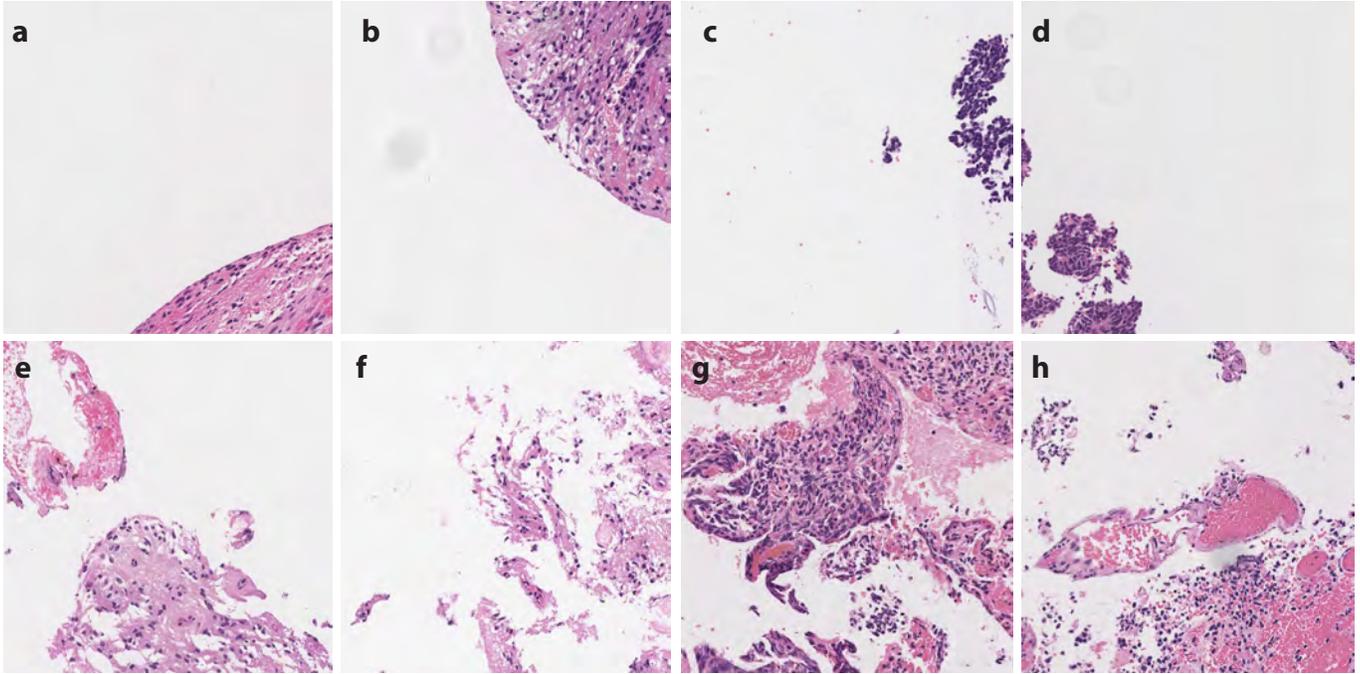

**Supplementary Fig. 4. The example patches from the independent test sets where the model predicted correctly with low confidence values.** Low confidence predictions were common in tumor images with significant white space as is seen for selected examples of ependymoma (a,b) or medulloblastoma (c, d). Samples with low confidence scores were also associated with tissue artifacts such as tissue chatter (e,f ependymoma) thermal artifact (g, medulloblastoma), or hemorrhage (h, medulloblastoma).

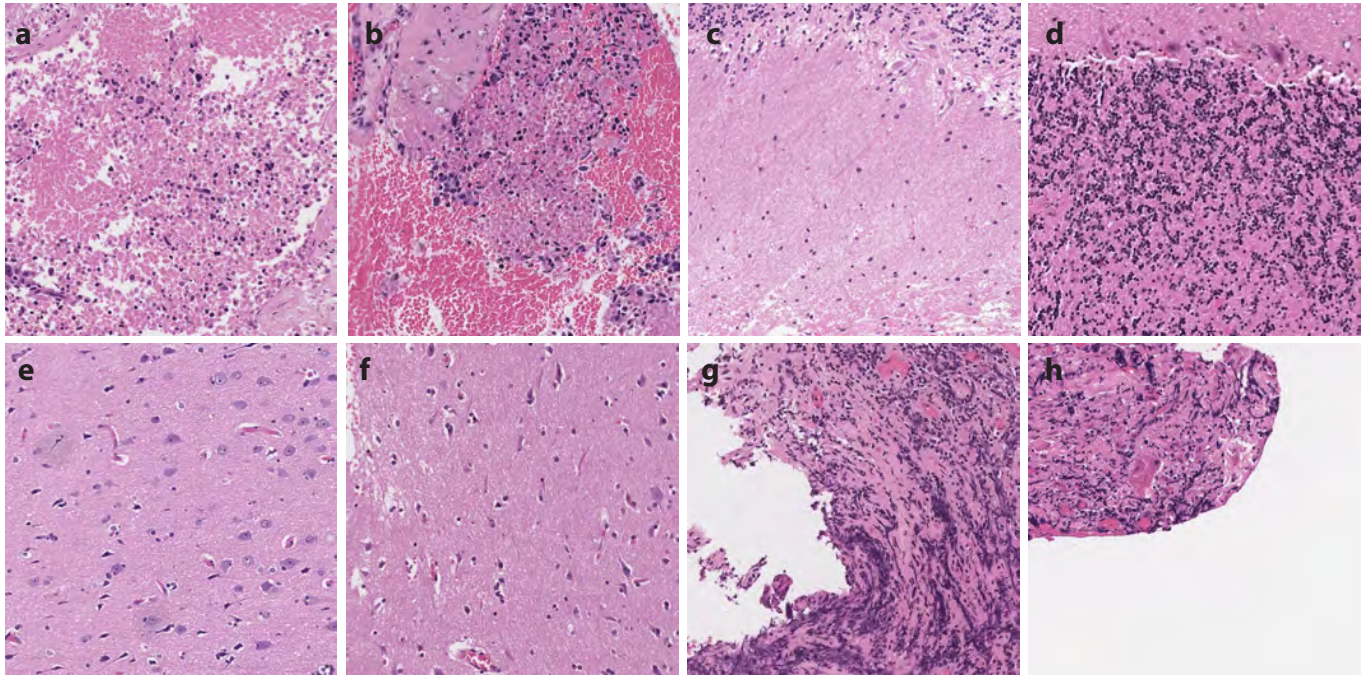

**Supplementary Fig. 5. Example patch images classified by the LRET-HFE trained ResNet50 model representing mislabeling of weakly annotated brain tumor dataset.** Selected patch images were correctly classified by the model and represent mislabeling of the weakly annotated brain tumor dataset. For instance, some tumors had ground truth of a tumor class such as a. breast adenocarcinoma or b. lymphoma, but were correctly classified as necrosis by the ResNet50-HFE trained model. Other images with ground truth labels representing a tumor class were correctly classified as normal tissue including c. a ground truth hemangioblastoma called cerebellum, d. ground truth pilocytic astrocytoma called cerebellum, e. ground truth white matter called grey matter, and f. pilocytic astrocytoma called grey matter. g, h. Examples of tumors with a ground truth label of lymphoma correctly classified as crush artifact by the model.